# Explainable predictions of different machine learning algorithms used to predict Early Stage diabetes.

**Vishal VAKIL[1], Shubh PACHCHIGAR[2], Chintan CHAVDA[1] and Sneh SONI[1,*]**

[1]Department of Instrumentation and Control Engineering, Institute of Technology, Nirma University, Gujarat, India
[2]Department of Electrical Engineering, Institute of Technology, Nirma University, Gujarat, India

(*Corresponding author's e-mail: snehsoni24@gmail.com)

**Running Title**
Diabetes prediction with explainable ML algorithms

**Abstract**

Machine Learning (ML) and Artificial Intelligence (AI) can be widely used to diagnose chronic diseases so that necessary precautionary treatment can be done in critical time. Diabetes Mellitus which is one of the major diseases can be easily diagnosed by several Machine Learning algorithms. Early stage diagnosis is crucial to prevent dangerous consequences. In this paper we have made a comparative analysis of several machine learning algorithms viz. Random Forest, Decision Tree, Artificial Neural Networks, K – Nearest Neighbor, Support Vector Machine and XGBoost along with feature attribution using SHAP to identify the most important feature in predicting the diabetes on a dataset collected from Sylhet Hospital. As per the experimental results obtained, the Random Forest algorithm has outperformed all the other algorithms with an accuracy of 99% on this particular dataset.

**Keywords:** Diabetes Mellitus, Machine Learning, K-Nearest Neighbor, Artificial Neural Networks, XGBoost, SHAP, Support Vector Machine, Decision Tree, Random Forest

**Introduction**

Diabetes Mellitus is considered one of the chief ailments due to which 34.2 million individuals have been affected in the US (10.5% of entire US citizens) [11]. Diabetes has affected the lifestyle of many individuals by increasing the medical expenses and health problems among them [12]. Not only has it created financial and health problems but it has also increased the death rates significantly. Such kind of disease has gathered greater attention from people and hence many researchers, government bodies and pharmaceutical companies have conducted research and study for the prevention of the disease by utilizing new technologies. Various machine learning algorithms have been developed to mimic the human brain. Utilizing such algorithms for the pre-diagnosis of diabetes can help us prevent the dangerous consequences of this fatal disease. There are two types of Diabetes:

1. Type 1 Diabetes - The body's healthy cells which produce insulin are being obliterated by the body's immune system and hence the body is unable to produce the necessary insulin [13].
2. Type 2 Diabetes - Here the body is able to produce the insulin though it is not able to utilize it properly [13].

Machine Learning (ML) algorithms such as Artificial Neural Networks (ANN), Logistic Regression, K - Nearest Neighbor (KNN), K - Means, Support Vector Machine (SVM), etc. are being used for predicting various medical maladies. ML algorithms are found effective for the pre-diagnosis of Diabetes Mellitus (DM).

**Literature Survey**

Thorough analysis carried out in [1] depicts that Artificial Intelligence in addition to data mining work are useful in the diagnosis and prediction of having diabetes, its complications and healthcare management. Diabetes is a metabolic malady where arbitrary blood glucose level blood glucose level prompts risk of a wide range of sicknesses like renal failure, heart disease, etc. Diabetes that is being diagnosed in [2], utilizes the KNN algorithm. Whereas, diabetes complications predicted in [3] are done by implementing ML technique towards data mining pipelines that could consolidate themselves conventional analytical plan, and withdraw information. A study which included the Pima Indians dataset using KNN, Decision Trees, Random Forest and SVM showed that among these four algorithms, the Decision tree was skillful to get to most elevated correctness of 73.82% [5]. However, following the removal of noisy data, the Random Forest as well as KNN (k=1 neighbor) were able to achieve an accuracy of 100%. But there isn't a single record of any further performance tests. Authors of [6] implemented three separate machine learning algorithms i.e., Naïve Bayes, SVM and Decision Trees. To their result they concluded that Naïve bayes have the peak accuracy among the three of 76.3%. Authors of [7] focused their research on a new approach and used a new dataset, however they used the Pima Indians for their reference unlike the others who used it for their results which is not totally acceptable because of the dataset being biased towards the female population. They implemented the ensemble method of ML using Practice Fusion dataset. And to their results they got a fairly enough accuracy of 89% on the Pima Indians dataset. And the Ensemble model on the Practice Fusion dataset yielded 85% accuracy.

**Dataset Analysis**

The dataset is extracted out of Sylhet Diabetes Hospital in Sylhet, Bangladesh. It was gathered using direct survey from the patients of Sylhet Diabetes Hospital and it was sanctioned by a doctor. The database contains samples of 520 patients, having 16 features in each sample. The features and its correlation with the next feature is shown in the **Figure 1** below:

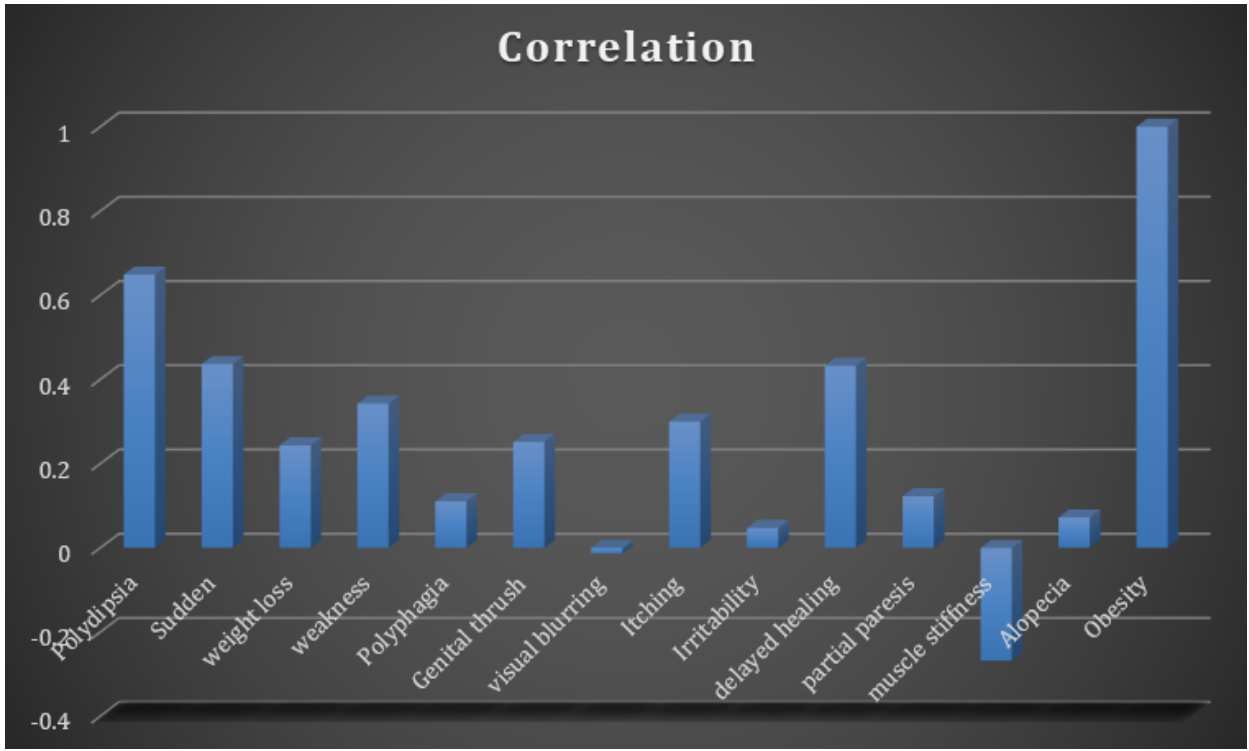

**Figure 1** Correlation of Features

**FEATURE ATTRIBUTION WITH SHAPLEY VALUES**

The field of eXplainable AI (XAI) that has emerged onto the scene in the last few years, aims to address the problem of interpretability. XAI researchers aim to define the very limits of interpretability, propose valid metrics, and craft novel algorithms that can allow us to peek at the inner workings of ML models, even if only partially. One of the most popular of these approaches involves using Shapley values (SVs) to attribute importance to features [17, 18]. Concretely, they originated with the motive on resolving the succeeding case—a class of distinctively expert member are all coordinating with one another for a common honor; Thus a forecast may be clarified by expecting because each characteristic value of the occasion is a "player" in a game where the extrapolation is the payout. Shapley values – one technique from alliance game hypothesis – discloses ourselves on how to reasonably disseminate "payout" among the quality. The clever trick is to leverage the game theoretical principles to give answers of feature attributions in machine learning. Hence, for a solitary illustration of the dataset, the "game" is the forecast task. If we subtract the average prediction for all instances from the real forecast for this instance, we get the "gain". The feature values of the instance i.e. the "players", associate to receive the gain (= predict a certain value).Here, S in val function for players defines the Shapley value. [16] The feature value has a Shapley value which is its input for payout, weighted and totalized for every realizable feature values fusion:

$$\delta_j(val) = \sum_{H \subseteq \frac{\{x_1, \ldots, x_i\}}{\{x_j\}}} \frac{|H|!(i-|H|-1)!}{i!} (val(H \cup \{x_j\}) - val(H))$$

Where $H$ is the subset of the features that are used in the model, $\delta$ is the Shapley value, $x$ represents the vector of feature values that are used to describe the instance and $i$ is the number of features [16]. In this paper we apply the principles of Shapley values and find the contribution of each feature towards the final output for each of our algorithms. We use the SHAP library to avoid the complex mathematical underpinnings and focus on the problem at hand.

- Xgboost Feature Attribution:

The graph below shows the features that contribute to the XgB model's output out of the base value ( aggregate output of model over the training dataset that we passed) on the output of model. The features that are in red are the one which make the prediction higher , while the one in blue are responsible for lower prediction. The plot shown within **Figure 2** clearly depicts that XgB model considers Polyuria to be the highest contributor to the final output, followed by Polydipsia.

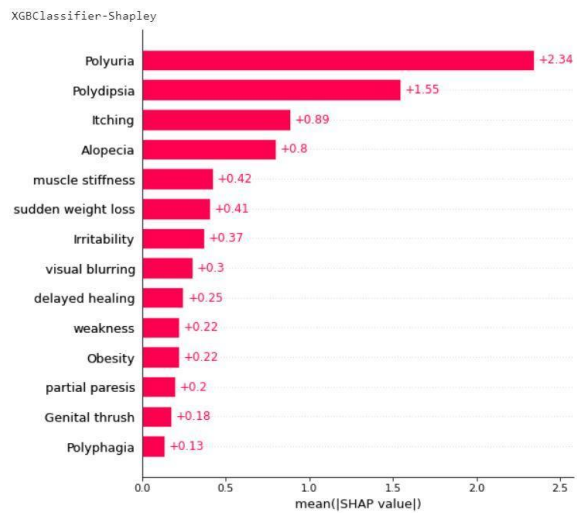

**Figure 2** Feature attribution for XG Boost

- SVM Feature Attribution:

A different way to picture the mentioned description is by applying force plot. The feature explanation bar has blue features that impel the risk of diabetes higher and red features that impel the risk lower. Each feature is categorized into its various groups based on the magnitude of its impact. From **Figure 3** it is observed that SVM construes Polyuria to be the most important feature as it has the greatest effect on final outcome. Similarly the presence of polyphagia has the least impact.

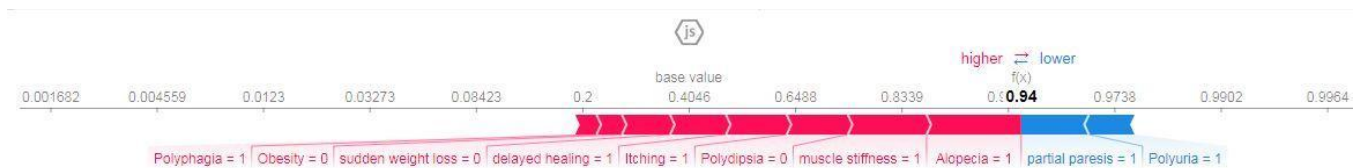

**Figure 3** SVM Feature attribution

- KNN Feature attribution:

    Force explanation plots obtained for KNN shown in **Figure 4** reveal the underlying feature attribution that KNN learns. Surprisingly, KNN considers partial paresis as the largest contributor while it does show the positive impact of the presence of polyuria. The presence of polyphagia has a negative impact on the overall risk of diabetes.

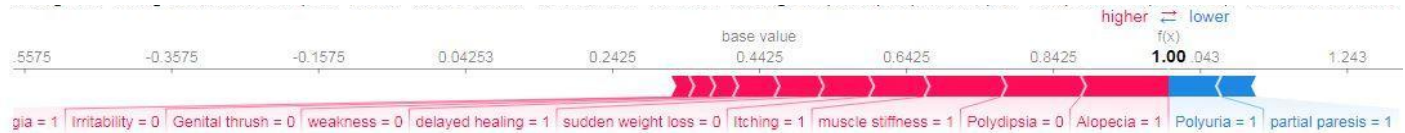

**Figure 4** KNN Feature attribution

- Random Forest feature attribution:

    A similar bar explainer plot is obtained for the Random Forest algorithm. The only change here is that the contribution factor is calculated for both classes i.e. in presence of as well as in the absence of particular feature. Even this algorithm attributes Polyuria with the highest mean Shapley value while considering obesity to have the least impact.

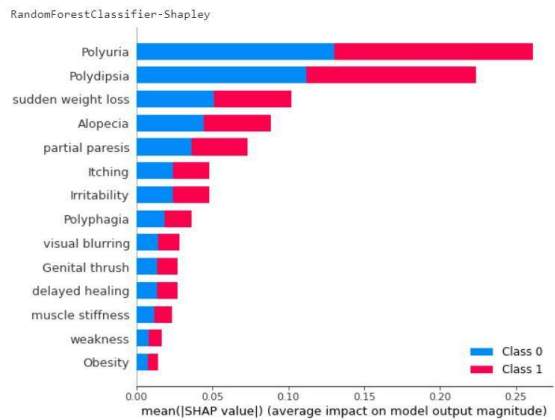

**Figure 5** Random Forest Feature attribution

- Artificial Neural Network Feature Attribution:

    The feature attribution plot for ANN is comparatively trickier to find, the crucial reason behind this is the presence of inherent correlations. Every neuron in an artificial neural network is depended on every other layer in a non-linear fashion. Non-Linearity is introduced by the SoftPlus activation [19]. Thus to disentangle all the layer weights from each other, Singular Value Decomposition (SVD) was implemented by the SHAP library. The SVD would then produce orthonormal weight vectors which are independent of perturbation in other layers; this lets us study them individually. Even ANN implicitly learns that Polyuria is the chief factor while predicting risk regarding diabetes.

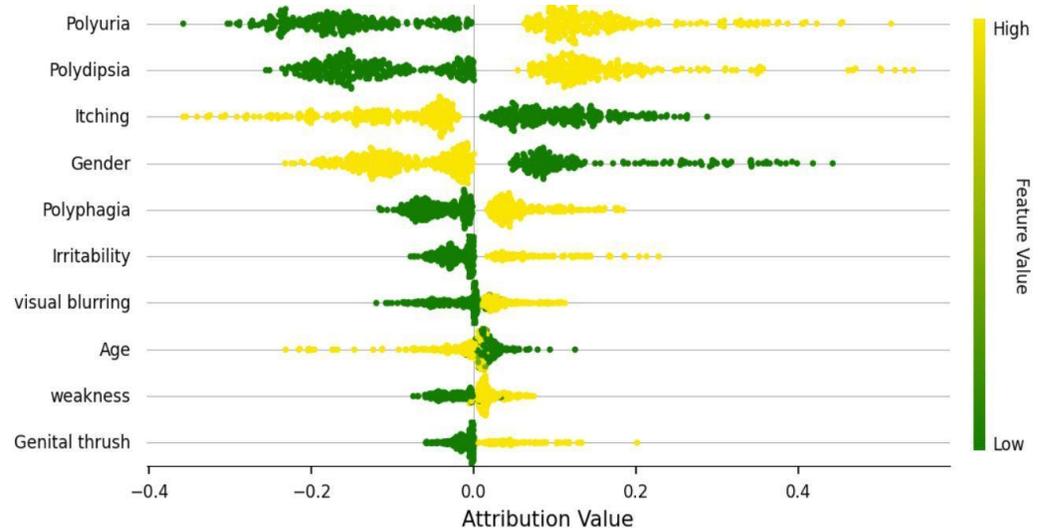

**Figure 6** ANN Feature attribution

**Algorithms Applied**

After thorough literature survey we found few algorithms suitable for early stage diabetes prediction. We implemented Decision Tree, Random Forest, K-Nearest Neighbor (KNN), Artificial Neural Networks (ANN), XGBoost algorithm and Support Vector Machine (SVM) on the Sylhet Hospital's dataset to analyze the best suitable algorithm for the desired prediction.

1. **Decision Tree**

Decision Tree is a flowchart having a treelike structure which has different cases in each node and according to the input the next node is selected. Each node has some prediction score according to which the final prediction is made. An image of the decision tree developed for our prediction is shown below:

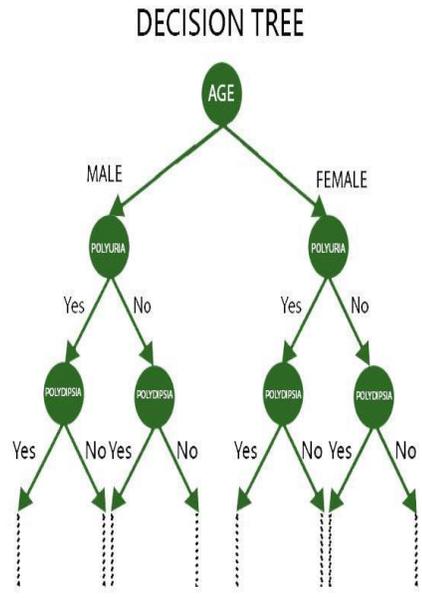

**Figure 7** Decision Tree developed for our dataset

This algorithm can be considered very effective for binary classification. Each feature will behave as a node and based on their value the prediction score will be calculated. Based on the importance of the feature the weightage is given to that node. Thus it evaluates each feature accurately and so the accuracy of this algorithm is higher for binary classification.

2. **Random Forest**

Random Forest is nothing but the group of different decision trees. So when any decision tree gives a wrong prediction then there are several others to predict correctly. The majority of the predicted class is considered as the final prediction. Thus it can provide the best accuracy. The below figure gives better idea to how the random forest algorithm works:

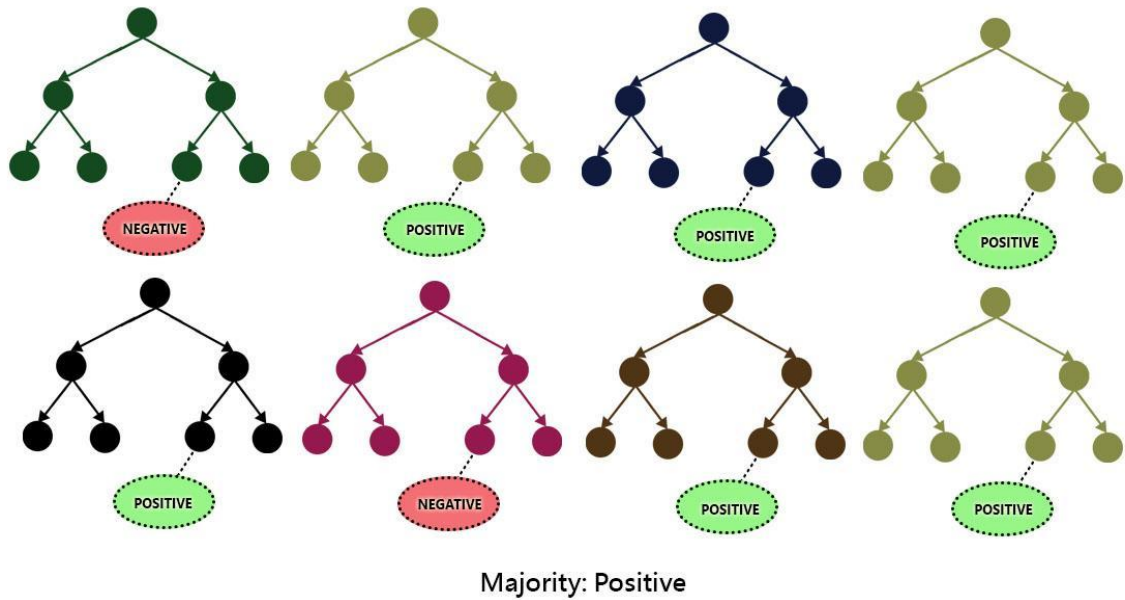

**Figure 8** Random Forest algorithm

As the majority of the decision trees predict Positive, the final prediction will be positive. This algorithm overcomes the issue of deep decision trees where trees that are grown very deep will in general adapt exceptionally sporadic examples: they over-fit their training sets, i.e. have low bias, yet extremely high variance [8].

3. **Artificial Neural Networks (ANN):**

ANN are analogous to the human brain where neurons interpret and pass information. ANN tries to mimic biological brain where the information i.e. Input is interpreted and then the processed information is passed ahead to different units or nodes and finally the output is generated. Each unit or node has some weight associated with it. According to the training dataset the weights are determined. At first they are randomly initialized and on iterations they are automatically adjusted to proper values to generate the desired output. The units or nodes of the ANN work just the same as neurons of a biological brain and each connection of those artificial neurons i.e. nodes acts like the synapses in a biological brain. Thus according to these connections of such artificial neurons the signal is processed and sent. Based on the weights of the respective nodes the strength of the signal is determined [9]. The artificial neural networks can have a single layer or multiple layers where multiple nodes are arranged in a layer. The "hidden layers" are those layers that are between the output and input layers, and the nodes that are organized in them are called "hidden nodes". ANN consists of both forward and backward propagations. The former involves the multiplication of weights with input and followed by addition of the bias to the product. Thereby applying the activation function to the input to propagate it forward. Whereas backward propagation, which is a crucial step in optimization of the model, finds the optimal parameters. An optimization function is required for backward propagation [14].

4. **K-Nearest Neighbor (KNN):**

For KNN, the item is classified according to the majority of nearest neighboring votes and can be assigned to class which is prevailing in its nearest neighbors. The value of K can be picked after plotting the graph of Error Rate VS K and find the appropriate errorless value. Also there are other techniques to select K. The downside of KNN is that recurring class favors the prediction of the new sample. However, we can overcome this by scaling the classification.

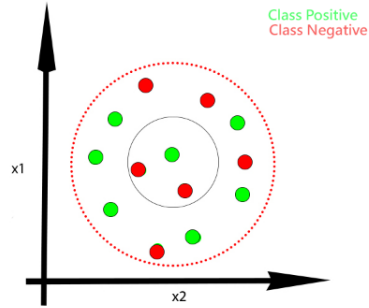

**Figure 9** Class Prediction through KNN

5. **Support Vector Machine (SVM):**

Support Vector Machine is a type of supervised learning algorithm that learns from the data collected by it and uses it to classify new samples. It builds a hyper plane used to classify the objects. The hyper plane has the largest distance possible to both the objects as higher the distance, lower the error of the classifier. For a larger dataset it can be hard to implement since it requires much time to train the data. SVM is better suitable for text recognition and to find the best linear separator. It is also hard to understand the final model of SVM [15]. To get tolerable results with SVM we can try to increase the features in the dataset which can help to increase the accuracy.

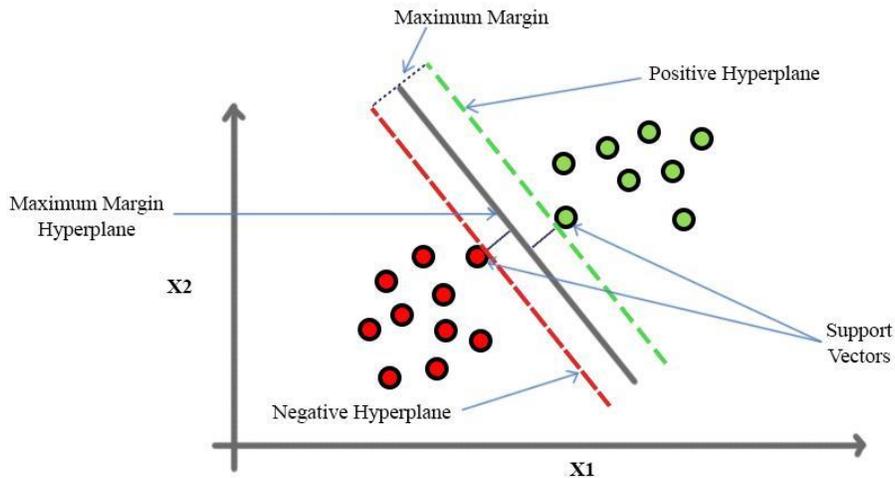

**Figure 10** SVM Algorithm

6. **XGBoost:**

XGBoost an ensemble ML algorithm which is a based on decision-tree, utilizes a gradient boosting framework. ANN tend to excel in the prediction problems which involves unstructured data (images, text, etc.), outperforming all other algorithms. However, the decision tree based algorithms yield best results during the prediction problems which involves small-to-medium structured/tabular data [4].

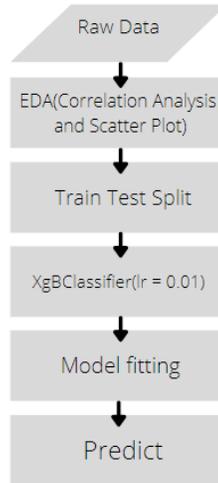

**Figure 11** XgB Algorithm based Diabetes Prediction

**Experimental Results**

We evaluate the proposed model on a dataset consisting of direct surveys conducted by a doctor on the patients of diabetes from Sylhet Diabetes Hospital hospital, Bangladesh. It is approved by a doctor which contains 16 features [10]. The database contains a total of 520 labeled examples in which the output is in binary form. After random shuffling of the data, we use the first 312 examples for training purposes. We implemented several algorithms to train the model with 250 epochs and learning rate of 0.30 in general. We also did a detailed comparative study on the respective outputs. **Table 1** lists the precision, recall and F1 score for the negative prediction of different algorithms implemented on the dataset. **Table 2** lists the precision, recall and F1 score for the positive prediction of different algorithms implemented on the dataset. **Table 3** lists the accuracy of different algorithms to predict exact results.

**Table 1** Precision, recall and F1-score for Negative prediction

| Algorithms | Precision | Recall | F1 Score |
|---|---|---|---|
| Random Forest | 0.98 | 0.98 | 0.98 |
| Decision Tree | 0.93 | 0.96 | 0.94 |
| ANN | 0.94 | 0.91 | 0.92 |
| KNN | 0.84 | 0.96 | 0.90 |
| SVM | 0.00 | 0.00 | 0.00 |
| XGBoost | 0.94 | 0.98 | 0.96 |

**Table 2** Precision, recall and F1-score for Positive prediction

| Algorithms | Precision | Recall | F1 Score |
|---|---|---|---|
| Random Forest | 0.99 | 0.99 | 0.99 |
| Decision Tree | 0.98 | 0.96 | 0.97 |
| ANN | 0.95 | 0.97 | 0.96 |
| KNN | 0.98 | 0.90 | 0.94 |
| SVM | 0.62 | 1.00 | 0.76 |
| XGBoost | 0.99 | 0.96 | 0.98 |

**Table 3** Accuracy of different algorithms:

| Algorithms | Accuracy |
|---|---|
| Random Forest | 0.99 |
| Decision Tree | 0.96 |
| ANN | 0.98 |
| KNN | 0.92 |
| SVM | 0.62 |
| XGBoost | 0.97 |

Decision tree has shown a great accuracy on the binary classification implemented on the mentioned dataset. The results we found from the implementation suggest that random forest has outperforms all the other five algorithms and shows the best accuracy as random forest is the collection of many decision trees where if one decision tree fails to provide the correct prediction then there would be another decision tree which would give the correct prediction. The class which is predicted by the majority of the decision trees turns out to be the final predicted class. Thereafter the Artificial Neural Networks (ANN) has also shown good performance on the model. While implementing ANN we had considered 1 input layer of size 16, 2 hidden layers - first of size 128 and second of size 64. There was one output layer of size 1. The data was shuffled randomly and then the maximum accuracy captured after three trials was noted. We found the accuracy of 95% through the present configuration of ANN.

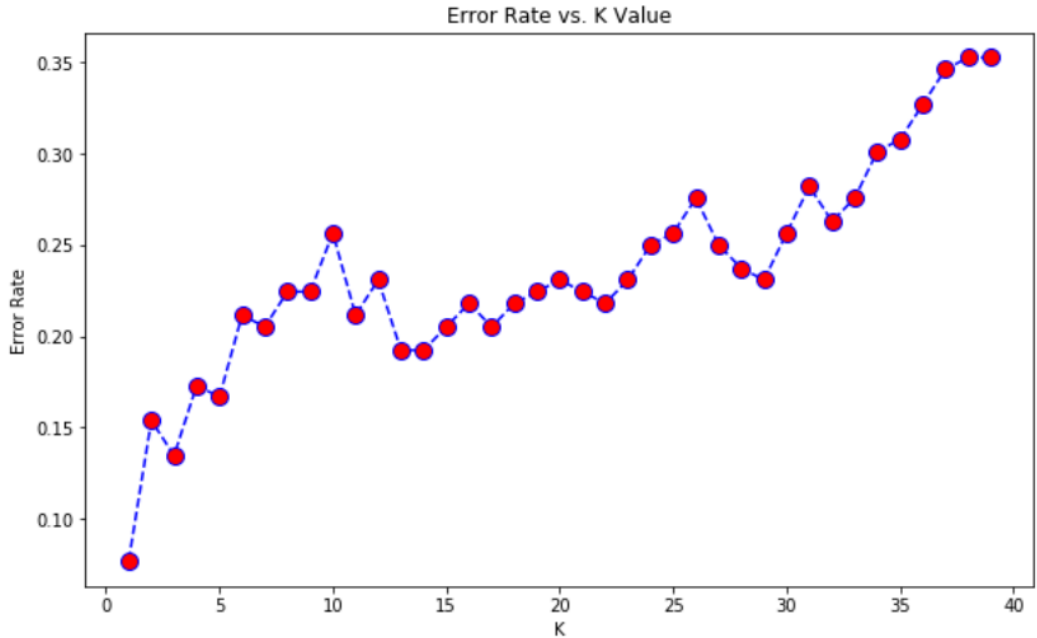

**Figure 12** Rate of Error vs K graph for KNN

For KNN, we plotted the rate or error v/s K graph as shown in the **Figure 12** and then selected the best value from the graph with the least rate of error. The SVM algorithm is usually used for classification. It is one of the most popular Supervised Learning algorithms which builds a model that predicts or classifies new examples to respective categories based on the marked training dataset. For this particular dataset SVM is not performing up to the mark as the dataset is somewhat biased in case of females. In such cases Random Forest performs very well as it considers all the features step by step and then predicts the class. For the XGBoost algorithm, the principle is carried out by iterative computation of weak classifiers. To get accurate and precise classification results, we propose to utilize XGBoost algorithm with data preprocessing in the in the forecast of diabetes with the experiment data from Sylhet Hospital establishing a classification model to anticipate the outcomes.

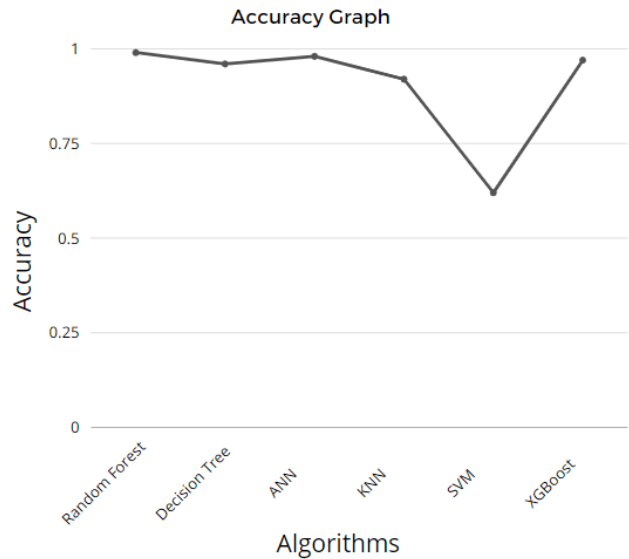

**Figure 13** Accuracy overview of all the algorithms

**CONCLUSION AND FUTURE WORK**

From the above algorithms implemented, we can observe that random forest has the best of accuracy in the prediction of diabetes with the experiment data from Sylhet Hospital's Dataset. Hence, we can recommend using this algorithm for the particular purpose. XGBoost has also outperformed all the algorithms except Random Forest by showing the accuracy of 0.97 i.e. similar to the Random Forest. Furthermore, other algorithms do their work up to their mark but SVM is not recommended as its results aren't tolerable. Decision Tree as compared to ANN, SVM and KNN has yielded good results with an accuracy of 0.96. The ANN also performs well on this dataset as compared to KNN and SVM. Though ANN possesses the accuracy of 0.95 which is less than Random Forest, XGBoost and Decision Tree, it can be recommended for the diabetes prediction. The KNN algorithm has also shown considerable results on the database. Feature attribution was also performed to find the most important feature of this dataset for diabetes prediction. We can see that Polyuria is the most important feature in the present dataset for predicting the diabetes among all the other features of this dataset. The least important feature in the diabetes prediction as compared to the other features of this dataset depends upon the algorithm applied as by applying the feature attribution on different algorithm yields different least contributing feature. In future a more comparative analysis can be done between different datasets and their features so that all the most important features can be identified for predicting the diabetes. Many different algorithms as well as combination of different algorithms can be tried to find the best and accurate diabetes prediction algorithm.